# Multilingual De-Duplication Strategies: Applying scalable similarity search with monolingual & multilingual embedding models

**Stefan Pasch**
BCG Plantinion
Frankfurt, Germany
stefan.pasch@outlook.com

**Dimitrios Petridis**
European Central Bank
Frankfurt, Germany
dimitrios.petridis@ecb.europa.eu

**Jannic Cutura**
European Central Bank &
DSTI School of Engineering
Frankfurt, Germany
jannic.cutura@dsti.institute

## Abstract

This paper addresses the deduplication of multilingual textual data using advanced NLP tools. We compare a two-step method involving translation to English followed by embedding with *mpnet*, and a multilingual embedding model (*distiluse*). The two-step approach achieved a higher F1 score (82% vs. 60%), particularly with less widely used languages, which can be increased up to 89% by leveraging expert rules based on domain knowledge. We also highlight limitations related to token length constraints and computational efficiency. Our methodology suggests improvements for future multilingual deduplication tasks.

## 1 Introduction

In this paper, we employ state-of-the-art natural language processing (NLP) tools to address the pervasive industry challenge of deduplicating multilingual textual data. This involves identifying pairs of text in different languages that convey the same content. Multilingual deduplication is vital in international and data-intensive business environments, where accurate analysis and consolidation of textual data across languages are required. In sectors such as job recruitment and e-commerce, identical content often appears in multiple languages. Effective deduplication is crucial for maintaining data integrity and ensuring that analyses reflect true and unbiased conditions.

Despite substantial progress in text deduplication methods, research focusing on multilingual contexts remains relatively limited. Current techniques predominantly address monolingual datasets, utilizing methods such as exact substring matching and suffix arrays to identify duplicate substrings efficiently (Lee & Ippolito, 2021). However, multilingual datasets introduce significant complexities, as simple word comparisons are inadequate across different languages. For example, the same job title may translate to "datastore manager" in one language and "database administrator" in another, with no word overlap but almost identical semantic content.

A common approach to capturing semantic similarity when there is no word overlap is to use text embeddings, which provide vector representations of texts (Reimers & Gurevych, 2019). Transformer-based embedding models are widely used in fields such as economics, finance, and organizational research (Tian & Pavur, 2023; Pasch & Petridis, 2023; Pasch & Ehnes, 2022; Lui et al., 2021; Koch & Pasch, 2023). However, these embedding models are typically trained on single-language corpora, and their capability to encapsulate and compare semantic meanings across different languages is inherently limited (Devlin et al., 2018).

To address this challenge, we apply and compare two approaches: (i) multilingual embedding models trained on datasets comprising multiple languages, which allows them to learn and understand semantic equivalences across these languages (Reimers & Gurevych, 2019), and (ii) translating all text inputs into English and then applying a monolingual English embedding model.

Determining the superior approach for multilingual deduplication is complex, as each strategy has distinct advantages and limitations. Multilingual embedding models are designed to understand and compare multiple languages directly, potentially reducing the loss of semantic nuances. However, these models may lack refinement or accuracy in specific languages due to the diversity and complexity of the data they are trained on. Conversely, translating all text into English to leverage a well-established English embedding model can capitalize on the advanced development and fine-tuning of these models. Nevertheless, this approach risks losing critical contextual and cultural nuances during the translation process, which can lead to errors in deduplication tasks.

To evaluate both approaches, we applied our methodology to a dataset provided by Eurostat as part of a data science competition. The dataset contains approximately 112,000 job advertisements, web-crawled from various online job portals. It includes multiple postings of the same job across different platforms, with variations in wording and language.

This dataset offers several unique features for assessing multilingual deduplication in a realistic industry setting: (i) It reflects the diverse nature of the European Union, encompassing job advertisements in 24 different languages, including typically underrepresented languages such as Lithuanian. (ii) Similar to typical industry applications, the dataset represents an unsupervised machine learning challenge where the true distribution of duplicates is unknown. Participants received automated feedback on a limited set of submissions, allowing benchmarking of different approaches while maintaining a real-world context. (iii) The dataset's size, with over 112,000 text entries, presents significant scalability challenges often encountered in industrial setups.

To address scalability, we utilized FAISS (Fast AI Similarity Search) index search, a scalable vector search engine (Johnson et al., 2019), in both of our approaches.

Our findings indicate that the two-step approach, involving translations and an English model, significantly outperforms the multilingual embedding model, achieving an F1 score of 82% compared to 60%. This supports the effectiveness of the two-step approach for multilingual deduplication in industry settings with the current state of embedding models.

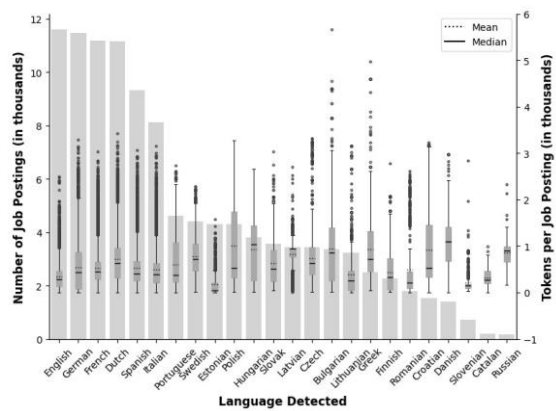

Figure 1: Distribution of languages and tokens per posting

Notably, the multilingual approach had an advantage in execution time, completing in approximately 1 hour on a single EC2 instance g4dn.xlarge, compared to 9 hours for the two-step approach (with 8 hours dedicated to translations, which could however be executed asynchronously). However, the perceived benefits of reduced execution time may vary based on hardware capabilities and specific time requirements.

Among 69 teams from 17 countries, our approach secured 2nd place in accuracy and 1st place in reproducibility. Replication files are available online[1].

## 2 Data

Our dataset, provided by Eurostat for the Web Intelligence Data Science Competition, comprises approximately 112,000 job advertisements sourced from online job portals such as LinkedIn and Xing. These postings include duplicates, meaning the same job may be described multiple times using different terminology or even different languages. The dataset includes approximately 24 languages, with the dominant ones being English, German, French, Dutch, Spanish, and Italian (see Figure 1). Each job posting includes details such as job title, company name, description, retrieval date, location, and a country identifier. However, web scraping often retrieves incomplete job postings,

---
[1] https://github.com/Dim10p/deduplication-challenge

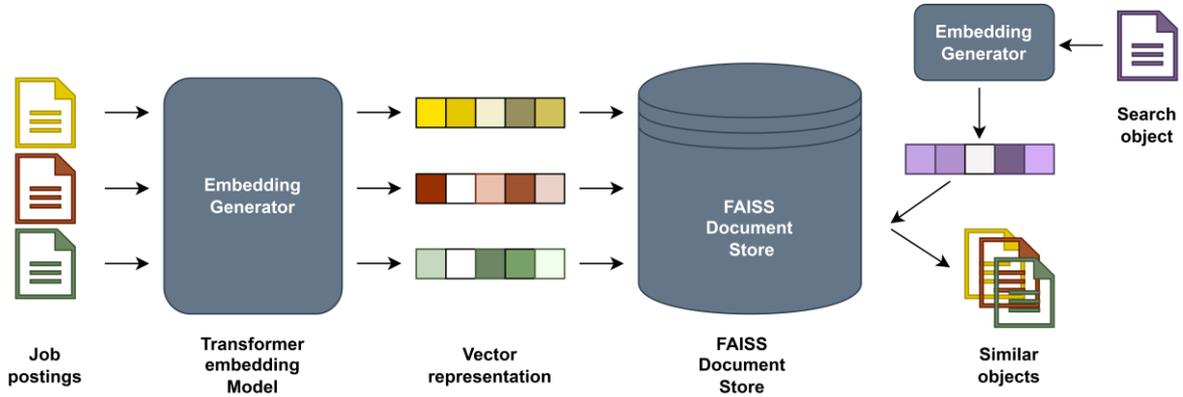

Figure 2: Methodology. Cleaned (and potentially translated) job postings are transformed using an embedding model, *mpnet* for English (Song et al. 2020) and *distiluse* (Reimers & Gurevych, 2019) for multi-language. The k-nearest-neighbors are retrieved using a Fast AI Similarity Search (FAISS) algorithm. Schematic representation adapted from Sumit (2024)

resulting in missing information. Specifically, company names are absent in about 25% of the observations, and location data is missing in roughly 50% of the observations, complicating the identification of patterns across job postings.

The competition organizers defined three types of duplicates to identify: (i) Full duplicates: job postings that are completely identical, having the same job title and description (up to minor discrepancies such as capitalization), but potentially differing in their sources and retrieval dates. (ii) Semantic duplicates: job postings that promote the same position and convey the same job characteristics (e.g., occupation, education, or qualification requirements) but are articulated differently in natural language or different languages. (iii) Temporal duplicates: either full or semantic duplicates that have different advertisement retrieval dates.

## 3 Methodology

### 3.1 Data preprocessing

We applied a minimal set of standard text cleaning procedures, including removing HTML tags from the web-scraping process (e.g., <br>, <strong>), converting all HTML character references to equivalent ASCII characters, retaining only ASCII characters and specific punctuations, removing unnecessary whitespace, splitting lowercase characters that precede uppercase characters, and removing repeated punctuations.

Following this initial preprocessing, we reduced the dataset to approximately 61,500 unique text inputs, effectively cutting the number of unique job postings by 50%. More importantly, this reduction decreased the number of pairwise comparisons from 6.2 billion to 1.9 billion, a reduction of 70%.

This rule-based text preprocessing enabled us to identify 99% of the full duplicates and accounted for 64% of the 73% of temporal duplicates that we were able to identify.

### 3.2 Embedding model and FAISS

A traditional approach to Natural Language Processing (NLP) is the dictionary method, which treats text inputs as a bag-of-words (Loughran & McDonald, 2011). For our deduplication task, this could involve measuring the overlap of words between two job advertisements. However, a significant challenge with our dataset is that many duplicates are not identical but instead have slightly different wording or are written in different languages. Additionally, translations may convert the same job title into different words, such as "manager" and "supervisor." Furthermore, with $n = 61,500$ (cleaned) job postings, this method would necessitate $\frac{n(n-1)}{2} = 1,89$ billion pairwise comparisons of two job titles and descriptions. Hence, the time complexity of a greedy algorithm which compares one by one is $O(n^2)$. To overcome these two challenges, our methodology relies on a combination of embedding models and Fast AI Similarity Search. Figure 2 visualizes the methodology.

**Embedding models** using transformers leverage the capabilities of advanced neural networks, such as BERT, GPT, or their successors, to transform text into dense vector representations. These embeddings capture the semantic nuances of words and sentences by considering their context within a broader text corpus. By training on large and diverse datasets, transformers learn to generate embeddings that encapsulate not just the surface meaning of words, but also their deeper connotations and relationships with other words. This makes embedding models particularly powerful for tasks like natural language understanding, text classification, and sentiment analysis, where understanding context is crucial (Devlin et al, 2018; Brown et al., 2020). For applications involving textual similarity, sentence transformers are regarded as the state-of-the-art method (Thakur et al, 2021). Different from Word-2-Vec (Rong, 2014), sentence transformers allow the generation of vector representation for longer text passages and not just single words. While different sentence transformer models exist, we rely on a model comparison from Reimers (2023) to apply the most performant English model, *all-mpnet-base-v2* (Song et al. 2020). For the non-translation approach, we use *distiluse* (Reimers & Gurevych, 2019).

**Fast AI Similarity Search** (FAISS) complements embedding models by providing a robust mechanism to quickly retrieve and rank similar embeddings from vast collections of text data. Using efficient indexing techniques and search algorithms, it enables real-time querying and comparison of embeddings, facilitating applications such as information retrieval, recommendation systems, and semantic search (John et al., 2021). When integrated with transformer-generated embeddings, Fast AI Similarity Search allows for highly accurate and swift identification of relevant information, enhancing the performance of AI-driven applications. This synergy between transformers and fast similarity search creates a powerful framework for handling large-scale text analysis and retrieval tasks with precision and speed (Lou et al, 2021).

**Translation to English** is performed using the Google Translate API, which is considered a suitable intermediate step for NLP tasks, even for languages that are semantically different from English (Ramadasa et al., 2022). In both our approaches (translate + embedding vs. multilingual embedding), the data is represented in vector form. FAISS efficiently retrieves the $k$ most similar job postings for each job post (with $k = 100$ for practical purposes) as candidate matches. This reduces the number of individual comparisons from 1.9 billion to 61.5 million, a reduction of over 99%. Algorithmically, this converts the problem to a linear time problem, given that translation, embedding, and FAISS search are each bound by $O(n)$ respectively (or $O(\log(n))$ if using IVF indexes).

A candidate pair is considered a semantic duplicate if the Euclidean distance (L2) is below a certain threshold. We experimented with several L2 distance thresholds ranging from 0.1 to 0.45 and found that a threshold of 0.25 is appropriate. This threshold filters out the 397,000 most semantically similar pairs, representing the top 6% most similar pairs (see Figure 1Figure 3). The distribution of L2 distances shows a dense concentration in the 0.25-0.50 range, indicating that higher thresholds would yield too many semantically dissimilar non-duplicates.

**Applying Expert Rules:** After filtering based on L2 distance, we further filtered potential duplicates using expert rules tailored to our dataset. For example, we applied varying thresholds depending on whether company name, language of the text, and job location were identical between two job postings.

## 4 Results

Since we are dealing with an unsupervised machine learning problem, creating labeled data for even a subset of observations is prohibitively expensive, making the evaluation of our approach challenging. Fortunately, the Eurostat Web Intelligence Data Science Challenge submission system provides F1 scores for each of a maximum of 10 submissions, allowing us to assess our model's performance. F1 scores are calculated based on whether each text pair is a Full, Semantic, Temporal, or No-Duplicate.

|  | Duplicates | | |
|---|---|---|---|
|  | **Full** | **Semantic** | **Temporal** |
| **Multilingual Model** | 0.99 | 0.60 | 0.57 |
| **Translation & English Model** | 0.99 | 0.82 | 0.67 |
| **Translation, English Model & domain expertise** | 0.99 | 0.89 | 0.92 |

Table 1: Detailed evaluation metrics. The value of each cell displays the F1 score for the three different types of duplicates.

Table 1 and Table 2 present a comprehensive overview of our evaluation. Two key findings emerge. First, the two-step approach yields a considerably higher F1 score of 82%, compared to 60% for the multilingual model[2]. Second, the two-step approach requires substantially more time to run (9 hours versus 1 hour). However, the additional 8 hours of runtime is due to the sequential Google Translate API calls. In a production-grade setup, these API calls could be executed asynchronously to reduce the runtime.

These results suggest that, despite the longer processing time, the two-step approach is more effective for multilingual deduplication with the current state of embedding models.

Further in the third panel of Table 1 we apply additional expert rules on the approach that appeared to be superior in extracting potential duplicate pairs (translations + English model). These expert rules based on domain knowledge helped us to further increase our F1 score to 89%.

## 5 Limitations

An important limitation of our approach is the token length constraint inherent in transformer models. Our primary model, *mpnet*, supports a maximum of 384 tokens, where a token roughly corresponds to a single word, although longer words may span two or three tokens. In practice, about one-third of our observations contain texts exceeding this 384-token limit. Consequently, any information beyond the 384th token is truncated and lost, which can be a significant limitation in applications involving lengthy texts. This truncation can result in the loss of critical context and details, potentially impacting the model's accuracy and effectiveness.

|  | Runtime | F1 score |
|---|---|---|
| **Translation and English Model** | ~9 hours | 82% |
| **Multilingual Model** | ~1 hour | 60% |

Table 2: Runtime vs. Accuracy

In our specific case, out of the 112,000 job advertisements analyzed, approximately 48,000 (43%) were truncated due to exceeding the 384-token limit. This truncation resulted in an average (median) loss of 540 (400) tokens per job posting for the affected entries. Figure 3 (a) illustrates the distribution of tokens per job advertisement, starting from the cutoff point imposed by this limitation.

Another limitation is that we currently evaluate only the 100 nearest neighbors for each observation before calculating the L2 distance, instead of directly limiting matches by L2 distance ("two-step approach"). However, as depicted in Figure 3 (b), this would only affect 881 job advertisements. For these, all 100 nearest matches have an L2 distance

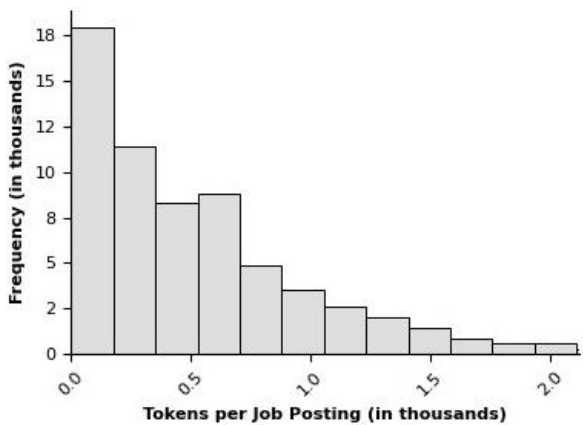

(a) Truncation severity

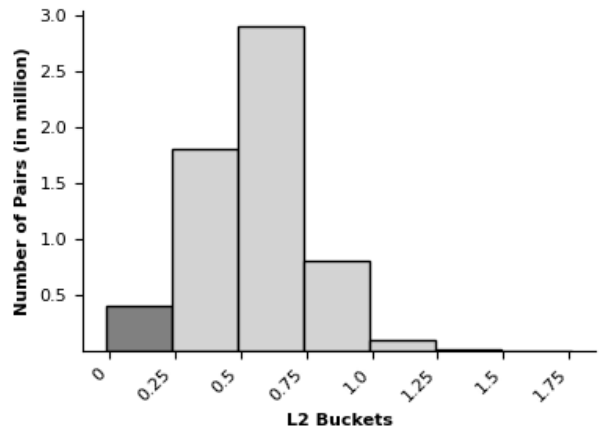

(b) Observations affected by two step filtering

Figure 3: Limitations of Embedding + FAISS

---

[2] We are considering the F1 score of semantic duplicates, since the full duplicates are not multilingual by construction, and the temporal ones are a combination of semantic and full duplicates.

below 0.25, indicating that the 101st match would likely have been relevant as well. In other words, the dark grey group of matched tuples would have been more numerous if we had not capped the potential matches to 100 before considering the L2 distance.

## 6 Discussion

Our study compared two approaches for multilingual deduplication of job advertisements: a two-step translation and English embedding method, and a direct multilingual embedding model. The two-step approach achieved a higher F1 score (82% vs. 60%), demonstrating its current effectiveness despite additional computational overhead.

While the multilingual model (*distiluse*) generated reasonable word embeddings for texts in different languages, our accuracy significantly improved with the translation step and the application of an "English-only" model (*mpnet*). Manual checks of a sub-sample revealed that the multilingual model struggled with less widely used languages. For instance, it falsely matched the Lithuanian job titles "pamainos vadvos" (shift manager) and "gamybos darboutojas – kokybes kontrolierius" (production worker - quality controller) as duplicates, highlighting its limitations.

A significant limitation is the token limit of transformer models like *mpnet*, which truncates texts exceeding 384 tokens, affecting 43% of our dataset. This truncation can lead to loss of critical context, impacting accuracy. Additionally, limiting evaluations to the top 100 nearest neighbors might exclude relevant matches, suggesting the need for more flexible neighbor selection methods.

The two-step approach's runtime of 9 hours, mainly due to sequential Google Translate API calls, highlights a trade-off between performance and efficiency. However, asynchronous processing could mitigate this in production settings.

Our methodology's effectiveness was validated by securing 2nd place in accuracy and 1st place in reproducibility in the Eurostat Web Intelligence Data Science Challenge, indicating its practical applicability.

Future work should focus on handling longer texts without truncation, improving neighbor selection flexibility, and optimizing translation processes to enhance both accuracy and efficiency in multilingual deduplication tasks.

## Acknowledgments

We gratefully acknowledge research support from DSTI School of Engineering.